\documentclass{article}
\usepackage{ijcai25}

\usepackage{times}
\usepackage{soul}
\usepackage{url}
\usepackage{etoolbox} % Needed for \forcsvlist

\usepackage[hidelinks]{hyperref}
\usepackage[utf8]{inputenc}
\usepackage[small]{caption}
\usepackage{graphicx}
\usepackage{amsmath, amssymb}
\usepackage{amsthm}
\usepackage{booktabs}
\usepackage{algorithm}
\usepackage{algorithmic}
\usepackage{xcolor}
\usepackage[switch]{lineno}
\usepackage{array}
\usepackage{booktabs}
\usepackage{enumitem}
\usepackage{hyperref}
\usepackage{tabularx}
\usepackage{tikz}

\usepackage{forest}
\usepackage{xcolor}
\usetikzlibrary{arrows.meta}

\definecolor{Apricot}{RGB}{251,185,130}
\definecolor{Mahogany}{RGB}{192,64,0}
\definecolor{JungleGreen}{RGB}{41,171,135}

\def\checkmark{\tikz\fill[scale=0.4](0,.35) -- (.25,0) -- (1,.7) -- (.25,.15) -- cycle;}

\newif\ifciteall
\citealltrue %change this to \citealltrue to always cite all references

\newcounter{partialcitecounter}

\newcommand{\partialcite}[2]{%
  \ifciteall
    \cite{#2}%
   \else
      \ifnum#1=0
      \else
        \setcounter{partialcitecounter}{0}%
        \def\citationlimit{#1}%
        \def\selectedcites{}%
        \forcsvlist{\appendtocite}{#2}%
        \cite{\selectedcites}%
      \fi % Closes the \ifnum block
  \fi
}

\newcommand{\appendtocite}[1]{%
  \stepcounter{partialcitecounter}%
  \ifnum\value{partialcitecounter}>\citationlimit
  \else
    \ifx\selectedcites\empty
      \edef\selectedcites{#1}%
    \else
      \edef\selectedcites{\selectedcites,#1}%
    \fi
  \fi
}
\title{A Survey on Large Language Models for Automated Planning}

\author{
Mohamed Aghzal$^1$
\and
Erion Plaku$^2$\thanks{{The work by E. Plaku is supported by (while serving at) the
National Science Foundation. Any opinion, findings, and conclusions or
recommendations expressed in this material are those of the authors and do
not necessarily reflect the views of the National Science Foundation.}}\and
Gregory J. Stein$^1$\And
Ziyu Yao$^1$\\
\affiliations
$^1$Department of Computer Science, George Mason University, Fairfax, Virginia, USA\\
$^2$National Science Foundation, Alexandria, Virginia, USA\\%}
\emails
\{maghzal, gjstein, ziyuyao\}@gmu.edu,
eplaku@nsf.gov
}

\begin{document}

\maketitle

\begin{abstract}

The planning ability of Large Language Models (LLMs) has garnered increasing attention in recent years due to their remarkable capacity for multi-step reasoning and their ability to generalize across a wide range of domains. While some researchers emphasize the potential of LLMs to perform complex planning tasks, others highlight significant limitations in their performance, particularly when these models are tasked with handling the intricacies of long-horizon reasoning. In this survey, we critically investigate existing research on the use of LLMs in automated planning, examining both their successes and shortcomings in detail. We illustrate that although LLMs are not well-suited to serve as standalone planners because of these limitations, they nonetheless present an enormous opportunity to enhance planning applications when combined with other approaches. Thus, we advocate for a balanced methodology that leverages the inherent flexibility and generalized knowledge of LLMs alongside the rigor and cost-effectiveness of traditional planning methods.  %We have provided a continually updated list of the papers surveyed in this study on the following repository: \href{https://github.com/MohamedAghzal/Awesome-LLM-Planning}{https://github.com/MohamedAghzal/Awesome-LLM-Planning}. 

\end{abstract}

\section{Introduction}

\begin{table*}[!ht]
\centering
\caption{Benchmarks for Planning and their Characteristics (Det.: Deterministic)} 

\label{tab:benchmarks}
\resizebox{\textwidth}{!}{
\begin{tabular}{p{3.5cm}lllccp{5.5cm}}
\toprule
\textbf{Dataset} & \textbf{Domain} & \textbf{Observ.} & \textbf{\# of Agents} & \textbf{Det.?} & \textbf{Horizon} & \textbf{Evaluation Metrics} \\
\toprule
\multicolumn{7}{c}{\textit{\textbf{Interactive Games}}} \\
\toprule
TextWorld [\citeyear{cote18textworld}] & Interactive Games & Partially & Single & $\times$ & Long & Game scores, \# of steps \\

Jericho [\citeyear{Hausknecht_Ammanabrolu_Côté_Yuan_2020}] & Interactive Games & Partially & Single & \checkmark & Long & Game scores \\
%\midrule

\toprule
\multicolumn{7}{c}{\textit{\textbf{Web and Computer Navigation}}} \\
\toprule

WebShop [\citeyear{yao22webshop}] & Web Navigation & Partially & Single & \checkmark & Long &  Success \\

WebArena [\citeyear{zhou2024webarena}] & Web Navigation & Partially & Single & \checkmark & Short & Success \\

Mind2Web [\citeyear{deng2024mind2web}] & Web Navigation & Partially & Single & \checkmark & Short & Success \\

OSWorld [\citeyear{xie2024osworld}] & Computer Navigation & Partially & Single & \checkmark & Short & Success \\
\toprule
\multicolumn{7}{c}{\textit{\textbf{Instruction Following and Task Planning}}} \\
\toprule
BabyAI [\citeyear{chevalier-boisvert2018babyai}] & Instruction Following & Partially & Single & \checkmark & Short & Success, Path Length \\

gSCAN [\citeyear{ruis2020gscan}] & Instruction Following & Fully & Single & \checkmark & Short & Exact Match Accuracy \\

AlfWorld [\citeyear{ALFWorld20}] & Household Tasks & Partially & Single & $\times$ & Long & Success \\

VirtualHome [\citeyear{Puig_2018_virtualhome}] & Household Tasks & Partially & Single & $\times$ & Long & Success \\

\toprule
\multicolumn{7}{c}{\textit{\textbf{Robotics and Physical Planning}}} \\
\toprule
SMART-LLM [\citeyear{kannan2024multiagent}] & Robot Task Planning & Partially & Multi & $\times$ & Short & Success, Task Completion Rate, Goal Condition Recall, Robot Utilization, Executability \\

RocoBench [\citeyear{zhao2024roco}] & Robot Task Planning & Partially & Multi & $\times$ & Short & Success, \# of Steps, Re-plan Attempts \\

PPNL [\citeyear{Aghzal2023CanLLM,aghzal2024look}] & Path Planning & Fully & Single & \checkmark & Long & Success, Optimality, Executability \\
Robotouille [\citeyear{gonzalez-pumariega2025robotouille}] & Asynchronous Instruction Following & Fully & Multi & \checkmark & Long & Success, Optimality \\
\toprule
\multicolumn{7}{c}{\textit{\textbf{Tool Use and API Integration}}} \\
\toprule
API-Bank [\citeyear{li2023apibank}] & Tool (API) Use & Partially & Single & \checkmark & Short & Correctness, Rouge \\

MINT [\citeyear{wang2024mint}] & Tool (API) Use & Partially & Single & \checkmark & Short & Success \\

%ToolHop [\citeyear{ye2025toolhopquerydrivenbenchmarkevaluating}] & Multi-hop QA & Partially & Single & \checkmark & Long & Correctness, Invocation Error \\

TravelPlanner [\citeyear{xie2024travelplanner}] & Travel Planning & Partially & Single & \checkmark & Short & Success \\

TaskBench [\citeyear{shen2024taskbenchbenchmarkinglargelanguage}] & Task Automation & Fully & Single & \checkmark & Short & Naturalness, Complexity, Alignment \\

\toprule
\multicolumn{7}{c}{\textit{\textbf{Multi-Domain}}} \\
\toprule
AgentBench [\citeyear{liu2024agentbench}] & Code, Games, Web & Partially & Multi & $\times$ & Long & Success, F1, Progress, Reward \\

PlanBench [\citeyear{valmeekam2023planbench}] & Blocksworld, Logistics & Fully & Single & \checkmark & Short & Correctness, Optimality, Verification \\

\toprule
\end{tabular}%
}
\vspace{-0.2in}

\end{table*}

{Planning, the process of devising a sequence of actions to achieve specific goals, serves as a cornerstone of intelligent behavior. This cognitive ability enables agents, both human and artificial, to navigate complex environments, adapt to changing circumstances, and anticipate future events. Recognizing that this skill is essential for intelligent behavior, automated planning has been a fundamental task in artificial intelligence since the field’s inception, playing a crucial role in enabling systems to reason about possible courses of action, optimize decision-making, and efficiently achieve desired outcomes across a wide range of applications.}

In this context, the role of large language models (LLMs) in planning has garnered increasing attention in recent years, though their limitations remain a topic of significant debate. The ``emergent'' capabilities of LLMs \partialcite{0}{wei2022emergentabilitieslargelanguage} initially sparked enthusiasm about their potential to function as standalone planners, with several methods demonstrating impressive planning abilities \cite{yao2023treethoughtsdeliberateproblem,hao2023reasoning}. However, subsequent research has scrutinized these claims, revealing key shortcomings \partialcite{1}{stechly2024chainthoughtlessnessanalysiscot,verma2024brittlefoundationsreactprompting}. In particular, while LLM agents show some promise in high-level short-horizon planning, they often fail to yield correct plans in long-horizon scenarios, where their performance can degrade significantly~\partialcite{2}{chen2024can,aghzal2024look}, rendering them impractical and unreliable. {Additionally, even in cases where they are successful, the costs of the plans they produce can be arbitrarily poor, a limitation that is often overlooked in the literature proposing the use of LLMs for planning-centric tasks.}

Despite these limitations, the general domain knowledge embedded in LLMs owing to large-scale pre-training offers valuable opportunities to enhance the flexibility of traditional planning systems. For instance, their ability to extract and interpret relevant contextual information from natural language allows these models to serve as interfaces for translating text into structured formal representations that can be seamlessly integrated with symbolic planners \cite{chen2024autotamp,zhang-etal-2024-pddlego}. Additionally, LLMs offer the potential to enrich planning systems with commonsense reasoning, bridging gaps in domain knowledge that conventional planners may struggle to address without the need for extensive manual engineering  \cite{zhang2023groundingclassicaltaskplanners}. Furthermore, as they are trained on large amounts of human-generated data, LLMs can implicitly encode human stylistic and qualitative preferences. Thus, LLMs can also function as evaluators, assessing plans based on qualitative and stylistic criteria that are often challenging to express explicitly \cite{guan2024tasksuccessenoughinvestigating}.

In this work, we present an overview of the literature on the integration of LLMs into automated planning, with an emphasis on long-horizon high-level planning applications. While the main focus of our work is on LLMs, the research we survey and the arguments we present are also applicable to LLMs augmented with a vision encoder, also known as vision-language models (VLMs). 
% We believe that a key oversight in the literature is insufficient attention to plan efficiency. While LLMs exhibit some limited capability in solving some planning tasks, the solutions produced are often suboptimal, rendering them impractical as standalone planners. Nevertheless, we argue that integrating LLMs' domain knowledge with traditional planning systems' precision would enhance the flexibility of planning systems. 
{We compare the state of research when using LLMs as planners and when integrating LLMs into traditional planning frameworks and argue that the latter presents a more flexible and promising solution.}
% {While previous surveys have explored this topic \cite{huang2024understandingplanningllmagents,Wang_2024}, their focus has been primarily on LLM-based autonomous agents more broadly, without specifically addressing long-horizon planning applications. We believe that narrowing the scope to this aspect is essential for analyzing both the potential and perils of applying LLMs in real-world planning applications, particularly in terms of the costs associated with generating and executing plans.}
While previous surveys have explored this topic, they focused on a broader scope of LLM-based autonomous agents \cite{huang2024understandingplanningllmagents,Wang_2024} or lack a systematic and in-depth discussion about the various uses of LLMs in planning and their limitations. \cite{li2024laspsurveyingstateoftheartlarge}
By narrowing our scope to long-horizon planning, our survey offers a deeper discussion of LLMs for planning and the potential as well as pitfalls for future research.

\section{Task Formulation and Benchmarks}

\subsection{Task Formulation}

A planning problem can be defined as a tuple \((S, A, T, s_\text{init}, G)\) where:
\begin{itemize}[noitemsep, leftmargin=*]
    \item \(S\) is a set of states representing all possible configurations of the environment.
    \item \(A\) is a set of actions available; in general, \(A\) can also be a function of the state.
    \item \(T: S \times A \to S\) is a transition function that maps a state and an action to a new state.
    \item \(s_\text{init} \in S\) is the initial state from which the  process begins.
    \item \(G \subseteq S\) is the set of goal states the task aims to achieve.
\end{itemize}

\noindent The objective is to find a sequence of actions \(\pi = (a_0, a_1, ..., a_{n-1})\), such that
\( s_{i+1} = T(s_i, a_i) \quad \text{for } i = 0, 1, \ldots, n-1, \)
where \( s_0 = s_\text{init} \) and \( s_n \in G \).

\subsection{Planning Benchmarks}

Table \ref{tab:benchmarks} presents a set of benchmark scenarios that have been used to evaluate the performance of LLMs in planning. We categorize the tasks addressed in each dataset according to the attributes below.

\noindent\textbf{Domain:} The application targeted in the benchmark refers to the high-level area or field where the planning agent operates. This attribute highlights the diverse range of tasks, from interactive games (e.g., \textbf{TextWorld} \cite{cote18textworld}, %\textbf{Jericho} \cite{Hausknecht_Ammanabrolu_Côté_Yuan_2020}) 
and classical planning tasks (e.g. path planning \cite{Aghzal2023CanLLM}, blocksworld \cite{valmeekam2023planbench}) to household task execution (\textbf{AlfWorld} \cite{ALFWorld20}), web navigation (\textbf{WebShop} \cite{yao22webshop}, \textbf{WebArena} \cite{zhou2024webarena}), or robot task planning (\textbf{SMART-LLM} \cite{kannan2024multiagent}, \textbf{RocoBench} \cite{zhao2024roco}). 

\paragraph{Observability:} Observability defines the extent to which the agent can access the true state \( s_t \in S \) of the environment at time \( t \). In a fully observable environment, the agent receives the complete state \( s_t \), as is the case in classical planning benchmarks such as \textbf{PlanBench} \cite{valmeekam2023planbench}. Conversely, in partially observable environments, the agent only receives observations \( o_t \in \mathcal{O} \), which provide incomplete information about the true state. This setting requires maintaining a belief state \( b_t \in \mathcal{B}_{S}\) over \( S \). This is the case in the tasks introduced in \textbf{BabyAI} \cite{chevalier-boisvert2018babyai} and \textbf{TextWorld} \cite{cote18textworld}. While fully observable environments allow direct reasoning over the true states \( \mathcal{S} \), partially observable settings require reasoning over the belief space \( \mathcal{B}_{S} \). {For instance, in BabyAI, the agent is tasked with exploring rooms within a grid to conduct certain tasks specified in natural language, but can only observe objects within its field of view. It must, thus maintain a memory of what has been encountered and form beliefs about the unrevealed information.}

\noindent\textbf{Number of Agents:} The number of agents involved in a benchmark reflects whether the environment requires single-agent or multi-agent planning. While single-agent tasks involve a solitary agent interacting with the environment to achieve its goals, multi-agent tasks require coordination, competition, or negotiation between multiple agents, necessitating strategies that account for other agents' actions and policies. This need for inter-agent interactions presents an additional challenge to benchmarks such as \textbf{SMART-LLM} \cite{kannan2024multiagent}.

\noindent\textbf{Deterministic vs. Stochastic:} This attribute characterizes the nature of transitions in the environment. In deterministic environments, the transition function \( T(s_t, a_t) = s_{t+1} \) produces a single next state \( s_{t+1} \) for every state-action pair \((s_t, a_t)\). Benchmarks like \textbf{PlanBench} \cite{valmeekam2023planbench} and \textbf{PPNL} \cite{Aghzal2023CanLLM} are examples of deterministic tasks. In contrast, stochastic environments introduce randomness, where \( T(s_t, a_t) \) defines a probability distribution over possible next states, introducing uncertainty to the agent's planning process. This is the case of several games in \textbf{TextWorld} \cite{cote18textworld} and %\textbf{Jericho} \cite{Hausknecht_Ammanabrolu_Côté_Yuan_2020}, 
\textbf{SMART-LLM} \cite{kannan2024multiagent}. {For example, in several games in \textbf{TextWorld}, the outcomes of some actions are non-deterministic: e.g., action \emph{go down} only has 75\% of moving the agent down. }

\noindent\textbf{Long-Horizon vs. Short-Horizon Planning:} The extent of foresight required for solving the task. Short-horizon planning involves tasks where the agent primarily plans for immediate or near-term goals, with a small number of steps before task completion, focusing on immediate feedback and short sequences of actions that are near-sighted and reactive. In contrast, long-horizon planning requires agents to make decisions considering distant outcomes, requiring it to account for long-term effects.

\noindent\textbf{Evaluation Metrics:}
% The evaluation criteria used to evaluate planning performance on the proposed benchmark. 
{Depending on the specific task demands, different datasets may evaluate an agent plan differently, as reflected in their evaluation metrics.}
While most datasets emphasize the success rate as the most crucial task, some use other metrics such as optimality and partial success. 

\noindent\textbf{\textit{The State of Planning Benchmarks}} The benchmarks discussed encompass a wide range of applications, providing valuable insights into task-specific capabilities. However, they also exhibit limitations that restrict their applicability and generalizability. Most benchmarks are designed within small, controlled environments that fail to capture the noise and uncertainty commonly encountered in real-world scenarios. Thus, strong performance on these benchmarks does not guarantee similar outcomes in novel or realistic settings. Moreover, the objectives of these tasks often emphasize success rates or task completion, neglecting crucial aspects such as the quality and cost of the plans.

\section{Approaches for using LLMs as Standalone Planners and Their Limitations}

\begin{figure}[tb]
    \centering
    \includegraphics[width=\linewidth]{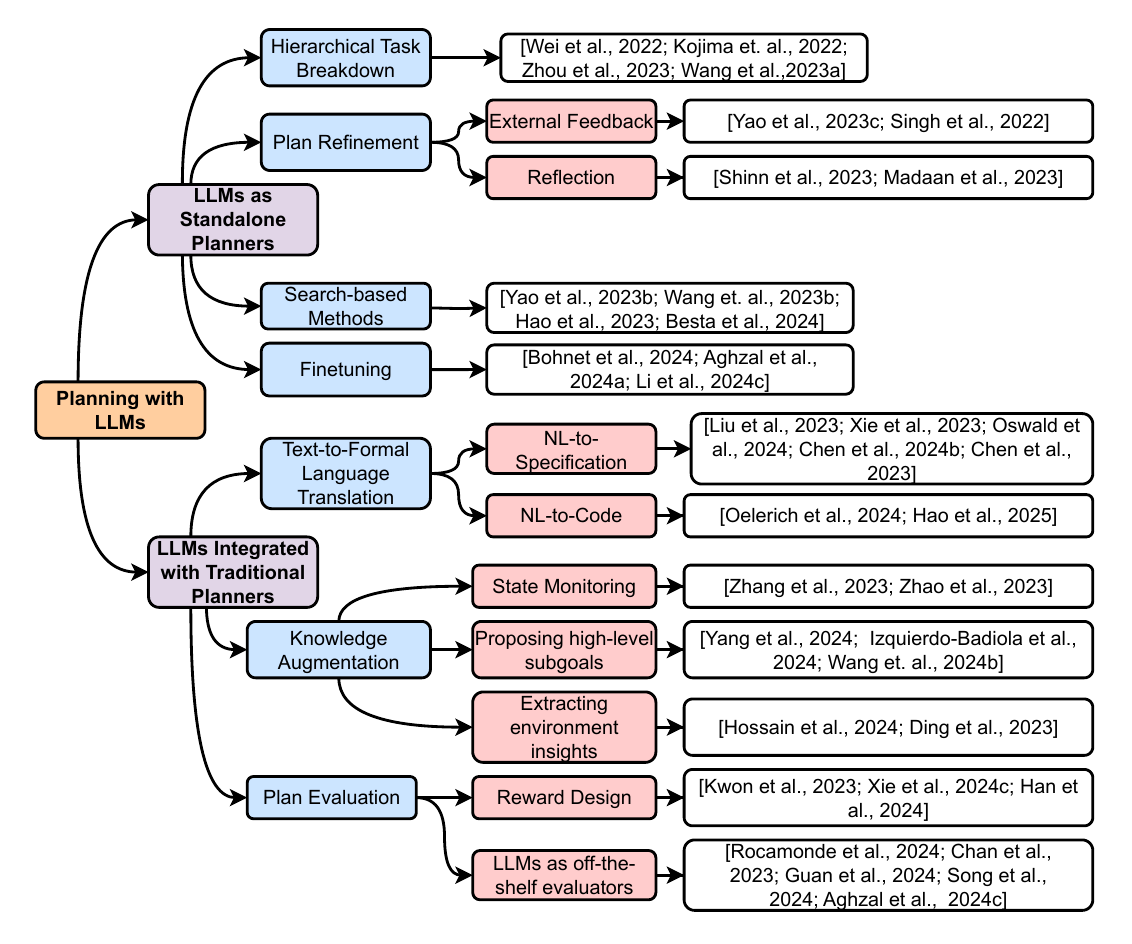}
        \vspace{-0.23in}
    \caption{Overview of Methods that Leverage LLMs for Planning and Decision-Making.}
    \label{fig:taxonomy}
        \vspace{-0.15in}
\end{figure}

% \begin{figure}[tb]
%     \centering
%     \includegraphics[width=\linewidth]{taxonomy-2.pdf}
%         \vspace{-0.3in}
%     \caption{Overview of Methods that Leverage LLMs for Planning and Decision-Making.}
%     \label{fig:taxonomy}
%         \vspace{-0.2in}
% \end{figure}

%\zyc{I wonder if the introduction of "LLM as planner" can be linked to the task formulations in Sec 2.} \gjsc{@Ziyu, I agree! Also, I don't know if the title of this section makes sense yet. You spend a bunch of this section just describing how LLMs plan. Perhaps, something more like ``Strategies for Planning using LLMs and Their Limitations''?}\zyc{Yes! I'm also not sure how much content we want to include here if it is only to highlight the limitations. However, I do think that we should have a paragraph clarifying how existing works have been using LLMs as planners (vs. how LLMs are used to assist planning, similarly another paragraph in the next section). Both paragraphs should link to the task formulation.}

% {\textbf{LLMs as end-to-end planners:} 
When using an LLM to directly propose plans, the state \( s \) is represented captured by the context provided in the input prompt, which encodes the environment, goals, and constraints. The LLM generates text or symbolic outputs conditioned on this input, effectively modeling the probability of selecting a plan \( \pi \) given the state \( s \). The initial state \( s_\text{init} \) is described in natural language: e.g., \emph{You are navigating a 7 by 7 maze. You are at location (3,3) and the goal is to reach location (6,5). Which direction do you choose to minimize steps to the goal?} \cite{Aghzal2023CanLLM}. The LLM's task is to implicitly simulate the subsequent states in order to identify the optimal policy. Since LLM reasoning occurs in the language space, it is crucial to restrict the action space to only executable options. This is often achieved by explicitly listing the possible actions (e.g., left, right, up, down) and providing examples of successful tasks to guide the model's outputs.
% , allowing the LLM to learn the allowable actions through few-shot demonstrations. 
In this section, we provide a categorization of common methods designed when using LLMs to produce plans and discuss the limitations of each.

\noindent\textbf{Hierarchical Task Breakdown:} 
A hierarchical task breakdown decomposes the planning problem  \( P = (S, A, T, s_\text{init}, G) \) into a high-level abstract planning problem  \( P^H = (S^H, A^H, T^H, s^H_\text{init}, G^H) \)  that produces an abstract plan \( \pi^H \) comprised of sub-tasks. Each abstract action in \( \pi^H \) is then refined into a detailed low-level plan for its corresponding subproblem, and integrating these refined plans yields the complete plan \( \pi \) that achieves the original goal. Leveraging LLMs for such decomposition involves breaking down complex tasks into smaller, manageable subtasks and solving each incrementally. This approach, exemplified by methods like Chain-of-Thought (CoT) \partialcite{2}{wei2022chain,kojima2022large}, Least-to-Most \cite{zhou2023leasttomost}, and Plan and Solve \cite{wang-etal-2023-plan}, operate under the assumption that while LLMs' may struggle with complex long-horizon tasks, they exhibit some proficiency in simpler short-horizon reasoning problems. 

\underline{\textit{Limitations:}} While promising, these techniques often depend on problem-specific, well-engineered prompts rather than robust algorithmic understanding \partialcite{1}{stechly2024chainthoughtlessnessanalysiscot}. LLMs struggle with tasks exceeding the complexity of the few-shot exemplars provided, limiting their practicality for long-horizon planning. Additionally, providing long reasoning sequences as demonstrations is constrained by LLMs' context limits and the associated performance degradation \cite{li2024longcontextllmsstrugglelong}. These techniques are also limited to cases where the environments are fully observable in order to be able to conduct this decomposition beforehand, however, this is rarely the case in real-world applications. Furthermore, while LLMs can be more successful at shorter horizon problems, the solutions produced are often far from optimal \cite{aghzal2024look}.  

\paragraph{Plan Refinement: } Another common approach in LLM-based planning involves iteratively improving plans based on feedback until achieving a goal. In this setting, the LLM continuously updates its beliefs regarding the environment and the effects of its actions, either through external information or by self-reflecting on its outputs in order to propose the plan. Several approaches \partialcite{2}{yao2023react,singh2022progprompt,sun2023adaplanner} decouple reasoning and action, allowing LLMs to propose a plan, execute it, and adjust using external feedback. Another direction leverages the idea that \emph{evaluation is easier than generation}, enabling LLMs to self-reflect and generate internal feedback for improvement \partialcite{2}{shinn2023reflexion,madaan2023selfrefine,kim2023rci}. 
These methods offer improved flexibility, as the iterative reasoning-and-acting allows the agent to dynamically update its strategy, making it particularly suitable in stochastic and partially observable settings.

\underline{\textit{Limitations:}}
However, feedback may not consistently improve strategies, as an LLM's ability to effectively utilize the feedback has proven to be limited \cite{verma2024brittlefoundationsreactprompting}. Additionally, there are no guarantees that the feedback information provided, especially in self-reflection, will be correct or useful, particularly in complex or domain-specific tasks \partialcite{1}{huang2024largelanguagemodelsselfcorrect,stechly2024selfverificationlimitationslargelanguage}. Moreover, the iterative nature of these methods can lead to inefficiencies, with performance degrading in tasks that require numerous iterations to reach a correct solution. It is possible for an LLM to succeed in a task by using an arbitrarily high amount of computation, by continuing to explore the space of possible plans. The high computational costs and slow inference make this approach impractical for long-horizon tasks where efficiency is paramount.

\paragraph{Search-Based Methods:} The idea that LLMs can effectively evaluate proposed solutions has also motivated their use in search-based methods for planning. In these methods, a state \( s \) encapsulates the current partial reasoning or problem-solving progress, the accumulated chain of thoughts and decisions made so far from which further reasoning (actions) can extend toward a complete solution. 
A basic example is \emph{CoT with self-consistency} (CoT-SC) \cite{wang2023selfconsistency}, which generates multiple reasoning chains and aggregates the results across all solutions. Building on this idea, more advanced methods like \emph{Tree of Thoughts} (ToT) \cite{yao2023tree} and \emph{Reasoning as Planning} (RAP) \cite{hao2023reasoning}
represent reasoning as an explicit tree structure, where nodes correspond to steps in the planning process, and branches are explored and evaluated to identify promising paths. In these methods, one LLM acts as a \emph{generator} to explore actions within the search space and another as a \emph{discriminator} to evaluate actions \cite{chen-etal-2024-tree}. Extensions like \emph{Graph of Thoughts} (GoT) \cite{besta2024got} introduce a graph-based structure to enable thought aggregation and refinement. 

\underline{\textit{Limitations:}} Despite their promise, these methods face significant limitations. The assumption that LLMs can serve as a sufficiently reliable model to evaluate the promise of actions is contentious. While LLMs encode some level of commonsense knowledge, their reasoning can falter in scenarios requiring them to predict the effects of an action several steps into the future. Additionally, the depth of the search tree or graph can grow exponentially for long-horizon tasks, leading to excessive computational and monetary costs that hinder their practicality in real-world applications. This is especially problematic in cases involving planning under uncertainty (e.g. partially observable settings), when the number of branches and possible scenarios can grow exponentially with every action. 
\paragraph{Finetuning:} Finetuning on sufficient amounts of data has proved to be effective in improving the quality of generated plans both in terms of correctness and efficiency \partialcite{3}{bohnet2024exploringbenchmarkingplanningcapabilities,Aghzal2023CanLLM,li2024unlockinglargelanguagemodels}. By exposing the model to a diverse and extensive dataset containing well-structured planning instances, it can infer patterns that approximate the policy used to produce the ground truth solutions and leads to success on similar problems. This is done by optimizing model parameters \(\theta\) so that the expected loss between the generated plan \( \pi \) and the ground-truth plan \( \pi^* \) over the training set \( \mathcal{D} \) is minimized: $\min_{\theta} \; \mathbb{E}_{(s_{init},\pi^*) \sim \mathcal{D}} \left[ \ell\left( \pi(s_\text{init}, \theta), \pi^* \right) \right]$. 
% for a given instruction and an input (initial) state \(s_\text{init}\).%: 

% \[
% \min_{\theta} \; \mathcal{L}(s_\text{init}, \theta) = \min_{\theta} \; \mathbb{E}_{\pi^* \sim \mathcal{D}} \left[ \ell\left( \pi(s_\text{init}, \theta), \pi^* \right) \right]
% \]

\underline{\textit{Limitations:}} However, these performance gains are limited to scenarios that closely resemble the training data, and the quality of the generated plans can drop dramatically by slightly changing parameters at test time. In practice, it is impossible to include every possible configuration of tasks and corresponding outputs in the training data, especially when dealing with long-horizon planning in stochastic and dynamic environments, which makes it difficult to account for all possibilities. Thus, fine-tuned models are unreliable unless the tasks encountered at inference time fall into the same distribution as the training data. This also makes deploying fine-tuned models as standalone planners data-inefficient, as the amount of data required to learn the necessary patterns can grow very large. 

\section{Integrating LLMs/VLMs with Traditional Planners}\label{sec:aided}

The limitations of LLMs in end-to-end planning have motivated the exploration of these models as auxiliary components in planning systems. The vast amounts of data fed to these models during pre-training allow them to serve as knowledge sources in a variety of applications. In this section, we divide research using LLMs as external components into three categories: 1) \emph{Text-to-Formal Specification Translation:} where an LLM plays the role of an interface between natural language specifications and formal languages that a symbolic planner can use; 2) \emph{Enhancing Planners with Commonsense Knowledge:} where LLMs serve as a heuristic to guide the planning process, allowing the planner to make decisions that are informed by external, commonsense knowledge, and 3) \emph{Plan Evaluation:} where LLMs serve the role of a critic by producing and refining reward functions or objective functions or by serving as off-the-shelf evaluators. We present a brief overview of these methodologies in Figure~\ref{fig:taxonomy}.

\paragraph{Text-to-Formal Language Translation:}
Building on the success of LLMs in machine translation and semantic parsing, several studies have explored their ability to translate task specifications expressed in natural language into a formal language that traditional symbolic planners can process to generate solutions. This approach effectively enables LLMs to serve as intuitive interfaces that allow tasks to be described naturally by users, while still leveraging the efficiency and rigor of established traditional planning frameworks.

To this end, multiple papers have investigated the use of LLMs for converting natural language specifications into formal languages such as the Planning Domain Definition Language (PDDL), which is then employed with classical planners to compute solutions \partialcite{3}{liu2023llmp,xie2023translatingnaturallanguageplanning,oswald2024large,guan2023leveraging,zhang-etal-2024-pddlego,smirnov2024generatingconsistentpddldomains}. In a similar vein, other studies \partialcite{2}{chen2024autotamp,chen-etal-2023-nl2tl,dainl2ltl} have focused on translating natural language instructions into temporal logic, offering a structured and formalized way to specify task constraints that must be met that can be used to guide the planning process. Others have also used LLMs as interfaces that translate natural language into code to interacts with traditional planning systems \partialcite{2}{oelerich2024languageguided,hao2025planning}.

Using LLMs in this context capitalizes on the advanced natural language understanding abilities of these models to simplify human-AI interaction, all while maintaining the advantages of traditional planners to generate high-quality and cost-efficient plans. Moreover, by allowing users to specify tasks directly in natural language, this approach makes planning systems significantly more accessible to non-experts who might otherwise be unfamiliar with formal languages.

%However, this flexibility can present itself as a double-edged sword; specifying tasks in natural language presents a number of ambiguities and complexities that are not present in formal languages. Thus, even if LLMs can serve as veridical translators, they might misunderstand user instructions or fail to capture subtle details, leading to incorrect or incomplete planning inputs. Additionally, running large LLMs can be computationally expensive, thus, for extensive or continuous planning needs, repeatedly invoking an LLM for instruction translation could become a bottleneck, and significantly slow down the planning process.

\noindent{}\paragraph{Enhancing Planners with Commonsense Knowledge:} The extensive scale and diversity of the data used to train LLMs equips them with exceptional levels of commonsense knowledge. This has motivated the integration of LLMs/VLMs into planning approaches, enabling the incorporation of domain-specific insights directly retrieved from these models. For instance, LLMs have been utilized to identify whether preconditions for specific actions are satisfied and to monitor the resulting effects of those actions \partialcite{2}{zhang2023groundingclassicaltaskplanners,zhao2023llmmcts}. Additionally, they can play a critical role in generating high-level sub-goals to guide long-horizon tasks \partialcite{3}{yang2024guidinglonghorizontaskmotion,Izquierdo2024plancolabnl,wang2024llm}. Another interesting application is their ability to provide statistical insights, such as estimating the likelihood of finding objects or the relationships between objects of 
interest in specific environments \partialcite{2}{hossain2024enhancing,ding2023task}, which can be particularly valuable in tasks like robotics and autonomous navigation. This approach draws on LLMs commonsense knowledge to relieve the need for manually engineering domain knowledge and supply semantic cues that guide the planner more efficiently than purely symbolic heuristics. This is also the driving motivation behind the LLM-modulo framework \cite{kambhampati2024position}, which advocates for leveraging the commonsense knowledge of LLMs to ``guess'' high-level plans and integrating them with external verifiers/checkers.

Such approaches are a promising use of LLMs, effectively allowing them to complement traditional planning algorithms with semantic heuristics and reasoning capabilities. This integration significantly enhances the flexibility of planning frameworks by providing strategic guidance that helps planners decompose complex tasks into structured sub-goals, without the need for manually encoding domain-specific knowledge. It can also help to reduce computational costs in long-horizon decision-making, by allowing LLMs to guide the search process based on the knowledge acquired during the pre-training stage. 

\noindent\textbf{Plan Evaluation:} Another promising direction involves using LLMs to evaluate plans proposed by traditional planners. In this approach, these models act as critics, providing feedback on proposed strategies. This is a particularly exciting application of LLMs. As they are exposed to large amounts of human-generated data during the pre-training process, LLMs implicitly learn desired stylistic considerations that are difficult to explicitly express in terms of an objective/reward function. Thus, these models can be well-suited as qualitative and stylistic plan evaluators.  

For example, several works use these models to design reward functions \partialcite{3}{kwon2023reward,xie2024textreward,han2024generatingevolvingrewardfunctions,li2023auto,corlRewardLLM}, where an LLM generates a reward function, which is then employed for RL policy learning. Alternatively, other research has explored directly using LLMs/VLMs as reward models \partialcite{2}{rocamonde2024visionlanguage,chan2023visionlanguage,zhong2024policy}. Other works leverage the commonsense reasoning capabilities of VLMs, employing them as scoring modules to detect undesirable behavior and provide feedback \partialcite{2}{guan2024tasksuccessenoughinvestigating,song2024vlmsocialnavsociallyawarerobot}. Furthermore, they can potentially be used to break the ties between different plans by considering human-like constraints, in applications where we are not only interested in plans that succeed but also those that are in accordance with human preferences \cite{aghzal2024evaluating}.

By incorporating LLMs/VLMs as evaluators, planning systems can be made significantly more versatile. Reward design is often a tedious process, and manually designed reward functions frequently fall short in capturing several important subtleties that can be described more naturally through language. Since LLMs are pre-trained on vast amounts of data, they implicitly encode many of these nuanced details, effectively recognizing complex patterns that traditional reward functions might overlook. Consequently, incorporating LLMs into planning algorithms as evaluators allows these systems to integrate commonsense knowledge directly into their reward and objective functions, enabling them to learn more efficiently and adaptively.

\section{Challenges and Opportunities}

Although LLMs provide significant potential when incorporated into planning systems, several challenges remain within this paradigm. In this section, we offer a detailed overview of the open problems, highlighting the areas that remain unresolved and that present opportunities for future work.

\paragraph{Addressing Ambiguity in Language to Task Formalization:} While LLMs can render planning significantly more flexible and enhance human-AI interaction by formalizing natural language specifications, this flexibility can also become a double-edged sword. The use of natural language to specify tasks introduces a host of ambiguities and complexities that are typically absent in formal languages, which are designed to be precise and unambiguous. Consequently, even when LLMs serve as veridical translators, there remains a risk that they may misunderstand user instructions or fail to capture the subtle details critical for accurate task representation, thereby leading to incorrect or incomplete inputs. Although the challenge of disambiguating instructions is a widely studied problem within the field of natural language processing \partialcite{0}{review2021ambiguities,niwa-iso-2024-ambignlg}, it is paramount to develop methods specifically tailored for planning languages. Such methods should be capable of efficiently handling scenarios that require the iterative invocation of an LLM not only to translate instructions but also to incorporate feedback, ensuring that the planning process remains reliable despite the inherent uncertainties of natural language input and the latencies introduced by the translation process.

\paragraph{Evaluating the Cost of Plans:} The cost of a plan, a numerical metric that quantifies the resources, time, or penalties associated with executing a particular sequence of actions, is an important consideration in planning. While LLMs have shown potential value across a range of applications, they have consistently exhibited notable limitations in numerical and metric reasoning \partialcite{1}{mirzadeh2024gsmsymbolicunderstandinglimitationsmathematical,ahn2024large}, a shortfall that restricts their ability to accurately assess, understand, and critique the costs associated with proposed plans. This inherent limitation not only diminishes the reliability of the cost evaluations provided by these models but also calls into question the robustness of the planning strategies that depend on such evaluations. As a consequence, enhancing the numerical and metric reasoning capabilities of LLMs emerges as a critical and promising direction for research, one that could substantially improve their effectiveness in guiding planning approaches toward truly optimal solutions. Consequently, finding ways to improve LLMs in this respect is a promising direction that could significantly increase their effectiveness in guiding planning approaches toward truly optimal solutions.

\paragraph{Improving Computational Efficiency:} While LLMs can be valuable tools for planning applications, they often come with significant computational overhead, as the monetary and computational costs of running inference on large models can severely limit their practicality, especially in applications requiring frequent queries (e.g. scenarios requiring iterative planning and repeated translation to formal specifications); these slow and expensive calls can create a bottleneck that reduces overall efficiency, and since some problems are inherently easier to solve than others, LLMs lack an inherent mechanism to distinguish between more difficult and easier tasks. Although approaches like ToT, OpenAI reasoning models\footnote{\href{https://openai.com/o1/}{{https://openai.com/o1/}}} and DeepSeek-R1\footnote{\href{https://ollama.com/library/deepseek-r1}{{https://ollama.com/library/deepseek-r1}}}\partialcite{0}{deepseekai2025deepseekr1incentivizingreasoningcapability} aim to bring these systems closer to solving complex reasoning tasks, they introduce excessive and often unnecessary computational overhead for tasks that could be handled more efficiently and can lead to significantly higher costs \partialcite{0}{valmeekam2024lrms}; thus, exploring methods for LLMs to balance answer quality and efficiency is an important endeavor. While finding methods to adapt which reasoning trajectories to task difficulty to improve efficiency is a topic that has garnered increasing interest \partialcite{2}{saha2025system,yue2025dots}, the direction remains in its infancy.   

\paragraph{Identifying and Mitigating Knowledge Gaps:} As we showcased in Section \ref{sec:aided}, the extensive knowledge acquired by LLMs can be highly valuable in guiding planners. Nevertheless, these models exhibit significant limitations when operating in highly specific domains or when faced with knowledge that is not well-represented in their training data, a shortcoming that often leads to hallucinations and non-factual outputs \cite{tonmoy2024comprehensivesurveyhallucinationmitigation}. In particular, LLMs can struggle in planning domains that demand reasoning about complex concepts and rules that were not encountered during training, thereby compromising their performance in such specialized areas. For instance, much of common sense knowledge, knowledge that emerges naturally from our direct interactions with the physical world, is not fully captured in text and thus remains largely absent from the data on which LLMs are trained \cite{lecun2022path}. Identifying these gaps in knowledge and developing methods to overcome them is an interesting research direction that could enhance the reliability of planning systems that leverage LLMs.

\paragraph{Interpretability and Explainability:} Another significant drawback of LLMs is that they exhibit a pronounced lack of interpretability, primarily because they learn to represent information through complex, high-dimensional structures that ultimately cause these models to operate as ``black boxes," thereby making it exceedingly difficult to discern the internal processes by which they make decisions. This lack of transparent decision-making presents challenges for the reliable and safe deployment of planning agents in real-world contexts that incorporate LLMs. For example, although LLMs hold the potential to enhance the versatility of planning frameworks through qualitative evaluation, the explanations they generate for their decisions often fail to accurately reflect their true internal reasoning processes, resulting in unfaithful explanations \partialcite{1}{faithfulreasoning,aghzal2024evaluating}. Addressing and investigating the causes of these failures may pave the way for the development of more reliable planning agents that incorporate LLMs. Additionally, a key unresolved question persists regarding what kinds of knowledge are actually encoded in LLMs. In other words, the precise nature of the information and implicit models incorporated within these systems remains largely unknown. This is an emerging area research \partialcite{1}{whatdollmsknow,vafa2024implicitworldmodels,li2023emergent}, however, contributions that specifically pertain to planning remain scarce.

\paragraph{Causal World Modeling:} As demonstrated by Richens and Everitt~[\citeyear{richens2024robust}], for robust agents to generalize to a wide variety of distributional shifts, they must learn an approximate causal world model that goes beyond merely capturing statistical correlations to understanding the true cause-and-effect relationships underlying their environments. The large-scale knowledge of LLMs holds the potential to enhance planning systems with causal reasoning based on commonsense knowledge. However, LLMs' capabilities in causal reasoning remain notably brittle~\cite{jin2024can}, highlighting a gap that can potentially be bridged by incorporating methods from causal inference into their training process, as advocated by Gupta et. al~[\citeyear{gupta2024essentialrolecausalityfoundation}]. While pre-training on large-scale observational data may impart a limited understanding of causal relationships, adapting to diverse scenarios requires the model to perform higher-level reasoning, such as interventions and counterfactual analysis, as outlined in Pearl's Causal Ladder \cite{bookofwhy}. This integration of causal inference techniques not only promises to enhance the robustness of these models under varying conditions but also represents an exciting research direction that could lead to the development of planning agents capable of more resilient and adaptable long-horizon decision-making.

\paragraph{Multi-Agent Planning:} Multi-agent planning tasks require multiple agents to work collaboratively in a coordinated manner to achieve a set of shared goals. Using LLMs as components to enhance multi-agent systems presents additional challenges. While inter-agent collaboration is essential, the inherent design of LLMs presents numerous challenges to their effective application in such settings\partialcite{0}{mutliagents2024p890}. In these scenarios, models are expected to manage a complex web of interactions, maintain clear and continuous communication, and respond adeptly to dynamic changes in the environment. This complexity necessitates that agents not only operate independently but also adapt their strategies in real-time based on the behaviors and decisions of other agents, a level of adaptability that current LLM-based agents have yet to achieve efficiently. The success of multi-agent collaboration is deeply rooted in robust communication and coordination between agents, as they must be able to share precise information about their goals, current states, and intended actions to collectively steer toward optimal outcomes. When current LLMs are used naively for such communication (e.g. by assigning an individual LLM to each agent), the result can be a significant increase in computational and monetary costs, thereby creating potential bottlenecks in system performance. Developing efficient communication methods that enable LLMs to coordinate effectively between different agents remains a largely under-explored research direction. 

\section{Conclusion}

In this work, we took a close look at the planning abilities of LLMs. We provided a set of benchmarks designed to evaluate these abilities and reviewed the various methodologies that have been proposed for incorporating LLMs into planning. We offered a critical examination of the performance of these techniques, assessing their effectiveness not only from the perspective of success rate but also in terms of cost. In addition, we highlighted some challenges and identified key areas for future research.

% \section*{Acknowledgments}
% %\ep{required for me}
% The work by E. Plaku is supported by (while serving at) the
% National Science Foundation. Any opinion, findings, and conclusions or
% recommendations expressed in this material are those of the authors and do
% not necessarily reflect the views of the National Science Foundation.


%\usepackage[maxnames=2]{biblatex}
\bibliographystyle{named}
\bibliography{ijcai25_truncated}

\begin{thebibliography}{}

\bibitem[\protect\citeauthoryear{Aghzal \bgroup \em et al.\egroup }{2024a}]{Aghzal2023CanLLM}
Mohamed Aghzal, Erion Plaku, and Ziyu Yao.
\newblock {Can Large Language Models be Good Path Planners? A Benchmark and Investigation on Spatial-temporal Reasoning}.
\newblock In {\em ICLR Workshop on LLM Agents}, 2024.

\bibitem[\protect\citeauthoryear{Aghzal \bgroup \em et al.\egroup }{2024b}]{aghzal2024look}
Mohamed Aghzal, Erion Plaku, and Ziyu Yao.
\newblock {Look Further Ahead: Testing the Limits of GPT-4 in Path Planning}.
\newblock {\em IEEE CASE}, 2024.

\bibitem[\protect\citeauthoryear{Aghzal \bgroup \em et al.\egroup }{2024c}]{aghzal2024evaluating}
Mohamed Aghzal, Xiang Yue, Erion Plaku, and Ziyu Yao.
\newblock Evaluating vision-language models as evaluators in path planning.
\newblock {\em arXiv:2411.18711}, 2024.

\bibitem[\protect\citeauthoryear{Ahn \bgroup \em et al.\egroup }{2024}]{ahn2024large}
Janice Ahn, Rishu Verma, et~al.
\newblock Large language models for mathematical reasoning: Progresses and challenges.
\newblock {\em arXiv:2402.00157}, 2024.

\bibitem[\protect\citeauthoryear{Besta \bgroup \em et al.\egroup }{2024}]{besta2024got}
Maciej Besta, Nils Blach, et~al.
\newblock {Graph of Thoughts: Solving Elaborate Problems with Large Language Models}.
\newblock {\em Proceedings of AAAI}, 2024.

\bibitem[\protect\citeauthoryear{Bohnet \bgroup \em et al.\egroup }{2024}]{bohnet2024exploringbenchmarkingplanningcapabilities}
Bernd Bohnet, Azade Nova, et~al.
\newblock Exploring and benchmarking the planning capabilities of large language models, 2024.

\bibitem[\protect\citeauthoryear{Chan \bgroup \em et al.\egroup }{2023}]{chan2023visionlanguage}
Harris Chan, Volodymyr Mnih, et~al.
\newblock {Vision-Language Models as a Source of Rewards}.
\newblock In {\em Second Agent Learning in Open-Endedness Workshop}, 2023.

\bibitem[\protect\citeauthoryear{Chen \bgroup \em et al.\egroup }{2023}]{chen-etal-2023-nl2tl}
Yongchao Chen, Rujul Gandhi, Yang Zhang, and Chuchu Fan.
\newblock {NL}2{TL}: Transforming natural languages to temporal logics using large language models.
\newblock In {\em EMNLP}, 2023.

\bibitem[\protect\citeauthoryear{Chen \bgroup \em et al.\egroup }{2024a}]{chen2024can}
Yanan Chen, Ali Pesaranghader, et~al.
\newblock Can we rely on {LLM} agents to draft long-horizon plans? {Let's} take {TravelPlanner} as an example.
\newblock {\em arXiv:2408.06318}, 2024.

\bibitem[\protect\citeauthoryear{Chen \bgroup \em et al.\egroup }{2024b}]{chen2024autotamp}
Yongchao Chen, Jacob Arkin, et~al.
\newblock {AutoTAMP}: Autoregressive task and motion planning with {LLM}s as translators and checkers.
\newblock In {\em IEEE ICRA}, 2024.

\bibitem[\protect\citeauthoryear{Chen \bgroup \em et al.\egroup }{2024c}]{chen-etal-2024-tree}
Ziru Chen, Michael White, Ray Mooney, Ali Payani, Yu~Su, and Huan Sun.
\newblock When is tree search useful for {LLM} planning? it depends on the discriminator.
\newblock In {\em Proceedings ACL (Volume 1: Long Papers)}, 2024.

\bibitem[\protect\citeauthoryear{Chevalier-Boisvert \bgroup \em et al.\egroup }{2019}]{chevalier-boisvert2018babyai}
Maxime Chevalier-Boisvert, Dzmitry Bahdanau, et~al.
\newblock Baby{AI}: First steps towards grounded language learning with a human in the loop.
\newblock In {\em ICLR}, 2019.

\bibitem[\protect\citeauthoryear{C\^ot\'e \bgroup \em et al.\egroup }{2018}]{cote18textworld}
Marc-Alexandre C\^ot\'e, \'Akos K\'ad\'ar, , et~al.
\newblock {TextWorld: A Learning Environment for Text-based Games}.
\newblock {\em Computer Games Workshop at ICML/IJCAI}, 2018.

\bibitem[\protect\citeauthoryear{Dai \bgroup \em et al.\egroup }{2024}]{dainl2ltl}
Zhirui Dai, Arash Asgharivaskasi, et~al.
\newblock {Optimal Scene Graph Planning with Large Language Model Guidance}.
\newblock In {\em IEEE ICRA}, 2024.

\bibitem[\protect\citeauthoryear{DeepSeek-AI}{2025}]{deepseekai2025deepseekr1incentivizingreasoningcapability}
DeepSeek-AI.
\newblock {DeepSeek-R1: Incentivizing Reasoning Capability in LLMs via Reinforcement Learning}, 2025.

\bibitem[\protect\citeauthoryear{Deng \bgroup \em et al.\egroup }{2024}]{deng2024mind2web}
Xiang Deng, Yu~Gu, et~al.
\newblock Mind2web: Towards a generalist agent for the web.
\newblock {\em NeurIPS}, 2024.

\bibitem[\protect\citeauthoryear{Ding \bgroup \em et al.\egroup }{2023}]{ding2023task}
Yan Ding, Xiaohan Zhang, et~al.
\newblock {Task and motion planning with large language models for object rearrangement}.
\newblock In {\em IEEE/RSJ IROS}. IEEE, 2023.

\bibitem[\protect\citeauthoryear{Gonzalez-Pumariega \bgroup \em et al.\egroup }{2025}]{gonzalez-pumariega2025robotouille}
Gonzalo Gonzalez-Pumariega, Leong~Su Yean, et~al.
\newblock {Robotouille: An Asynchronous Planning Benchmark for {LLM} Agents}.
\newblock In {\em ICLR}, 2025.

\bibitem[\protect\citeauthoryear{Guan \bgroup \em et al.\egroup }{2023}]{guan2023leveraging}
Lin Guan, Karthik Valmeekam, et~al.
\newblock {Leveraging Pre-trained Large Language Models to Construct and Utilize World Models for Model-based Task Planning}.
\newblock In {\em NeurIPS}, 2023.

\bibitem[\protect\citeauthoryear{Guan \bgroup \em et al.\egroup }{2024}]{guan2024tasksuccessenoughinvestigating}
Lin Guan, Yifan Zhou, et~al.
\newblock {Task Success is not Enough: Investigating the Use of Video-Language Models as Behavior Critics for Catching Undesirable Agent Behaviors}.
\newblock In {\em COLM}, 2024.

\bibitem[\protect\citeauthoryear{Guo \bgroup \em et al.\egroup }{2024}]{mutliagents2024p890}
Taicheng Guo, Xiuying Chen, et~al.
\newblock Large language model based multi-agents: A survey of progress and challenges.
\newblock In {\em IJCAI}, 2024.
\newblock Survey Track.

\bibitem[\protect\citeauthoryear{Gupta \bgroup \em et al.\egroup }{2024}]{gupta2024essentialrolecausalityfoundation}
Tarun Gupta, Wenbo Gong, et~al.
\newblock {The Essential Role of Causality in Foundation World Models for Embodied AI}.
\newblock {\em arXiv}, 2024.

\bibitem[\protect\citeauthoryear{Han \bgroup \em et al.\egroup }{2024}]{han2024generatingevolvingrewardfunctions}
Xu~Han, Qiannan Yang, et~al.
\newblock {Generating and Evolving Reward Functions for Highway Driving with Large Language Models}.
\newblock {\em arXiv}, 2024.

\bibitem[\protect\citeauthoryear{Hao \bgroup \em et al.\egroup }{2023}]{hao2023reasoning}
Shibo Hao, Yi~Gu, et~al.
\newblock Reasoning with language model is planning with world model.
\newblock In {\em EMNLP}, 2023.

\bibitem[\protect\citeauthoryear{Hao \bgroup \em et al.\egroup }{2025}]{hao2025planning}
Yilun Hao, Yang Zhang, et~al.
\newblock Planning anything with rigor: General-purpose zero-shot planning with {LLM}-based formalized programming.
\newblock In {\em ICLR}, 2025.

\bibitem[\protect\citeauthoryear{Hausknecht \bgroup \em et al.\egroup }{2020}]{Hausknecht_Ammanabrolu_Côté_Yuan_2020}
Matthew Hausknecht, Prithviraj Ammanabrolu, et~al.
\newblock Interactive fiction games: A colossal adventure.
\newblock {\em Proceedings of AAAI}, 2020.

\bibitem[\protect\citeauthoryear{Hossain \bgroup \em et al.\egroup }{2024}]{hossain2024enhancing}
Shahriar Hossain, Abhishek Paudel, and Gregory~J. Stein.
\newblock Enhancing object search by augmenting planning with predictions from large language models.
\newblock In {\em 2nd CoRL Workshop on Learning Effective Abstractions for Planning}, 2024.

\bibitem[\protect\citeauthoryear{Huang \bgroup \em et al.\egroup }{2024a}]{huang2024largelanguagemodelsselfcorrect}
Jie Huang, Xinyun Chen, et~al.
\newblock Large language models cannot self-correct reasoning yet.
\newblock In {\em ICLR}, 2024.

\bibitem[\protect\citeauthoryear{Huang \bgroup \em et al.\egroup }{2024b}]{huang2024understandingplanningllmagents}
Xu~Huang, Weiwen Liu, et~al.
\newblock {Understanding the planning of LLM agents: A survey}, 2024.

\bibitem[\protect\citeauthoryear{Izquierdo-Badiola \bgroup \em et al.\egroup }{2024}]{Izquierdo2024plancolabnl}
Silvia Izquierdo-Badiola, Gerard Canal, et~al.
\newblock {PlanCollabNL: Leveraging Large Language Models for Adaptive Plan Generation in Human-Robot Collaboration}.
\newblock In {\em ICRA}, 2024.

\bibitem[\protect\citeauthoryear{Jin \bgroup \em et al.\egroup }{2024}]{jin2024can}
Zhijing Jin, Jiarui Liu, et~al.
\newblock {Can Large Language Models Infer Causation from Correlation?}
\newblock In {\em ICLR}, 2024.

\bibitem[\protect\citeauthoryear{Kambhampati \bgroup \em et al.\egroup }{2024}]{kambhampati2024position}
Subbarao Kambhampati, Karthik Valmeekam, et~al.
\newblock {Position: {LLM}s Can{\textquoteright}t Plan, But Can Help Planning in {LLM}-Modulo Frameworks}.
\newblock In {\em ICML}, 2024.

\bibitem[\protect\citeauthoryear{Kannan \bgroup \em et al.\egroup }{2024}]{kannan2024multiagent}
Shyam~Sundar Kannan, Vishnunandan L.~N. Venkatesh, et~al.
\newblock {SMART-LLM: Smart Multi-Agent Robot Task Planning using Large Language Models}.
\newblock In {\em 2024 IEEE/RSJ IROS}, 2024.

\bibitem[\protect\citeauthoryear{Kim \bgroup \em et al.\egroup }{2024}]{kim2023rci}
Geunwoo Kim, Pierre Baldi, et~al.
\newblock {Language models can solve computer tasks}.
\newblock In {\em NeurIPS}, 2024.

\bibitem[\protect\citeauthoryear{Kojima \bgroup \em et al.\egroup }{2022}]{kojima2022large}
Takeshi Kojima, Shixiang~Shane Gu, et~al.
\newblock Large language models are zero-shot reasoners.
\newblock {\em NeurIPS}, 2022.

\bibitem[\protect\citeauthoryear{Kwon \bgroup \em et al.\egroup }{2023}]{kwon2023reward}
Minae Kwon, Sang~Michael Xie, et~al.
\newblock {Reward Design with Language Models}.
\newblock In {\em ICLR}, 2023.

\bibitem[\protect\citeauthoryear{LeCun}{2022}]{lecun2022path}
Yann LeCun.
\newblock A path towards autonomous machine intelligence.
\newblock {\em Open Review}, (1), 2022.

\bibitem[\protect\citeauthoryear{Li \bgroup \em et al.\egroup }{2023a}]{li2023emergent}
Kenneth Li, Aspen~K Hopkins, et~al.
\newblock Emergent world representations: Exploring a sequence model trained on a synthetic task.
\newblock In {\em ICLR}, 2023.

\bibitem[\protect\citeauthoryear{Li \bgroup \em et al.\egroup }{2023b}]{li2023apibank}
Minghao Li, Yingxiu Zhao, et~al.
\newblock {API}-bank: A comprehensive benchmark for tool-augmented {LLM}s.
\newblock In {\em EMNLP}, 2023.

\bibitem[\protect\citeauthoryear{Li \bgroup \em et al.\egroup }{2024a}]{li2023auto}
Hao Li, Xue Yang, et~al.
\newblock {Auto MC-Reward: Automated Dense Reward Design with Large Language Models for Minecraft}.
\newblock In {\em IEEE/CVF CVPR}, 2024.

\bibitem[\protect\citeauthoryear{Li \bgroup \em et al.\egroup }{2024b}]{li2024laspsurveyingstateoftheartlarge}
Haoming Li, Zhaoliang Chen, et~al.
\newblock {LASP: Surveying the State-of-the-Art in Large Language Model-Assisted AI Planning}, 2024.

\bibitem[\protect\citeauthoryear{Li \bgroup \em et al.\egroup }{2024c}]{li2024longcontextllmsstrugglelong}
Tianle Li, Ge~Zhang, et~al.
\newblock Long-context llms struggle with long in-context learning, 2024.

\bibitem[\protect\citeauthoryear{Li \bgroup \em et al.\egroup }{2024d}]{li2024unlockinglargelanguagemodels}
Wenjun Li, Changyu Chen, et~al.
\newblock {Unlocking Large Language Model's Planning Capabilities with Maximum Diversity Fine-tuning}, 2024.

\bibitem[\protect\citeauthoryear{Liu \bgroup \em et al.\egroup }{2023}]{liu2023llmp}
Bo~Liu, Yuqian Jiang, et~al.
\newblock {LLM+P: Empowering Large Language Models with Optimal Planning Proficiency}, 2023.

\bibitem[\protect\citeauthoryear{Liu \bgroup \em et al.\egroup }{2024}]{liu2024agentbench}
Xiao Liu, Hao Yu, et~al.
\newblock {AgentBench: Evaluating {LLM}s as Agents}.
\newblock In {\em ICLR}, 2024.

\bibitem[\protect\citeauthoryear{Madaan \bgroup \em et al.\egroup }{2023}]{madaan2023selfrefine}
Aman Madaan, Niket Tandon, et~al.
\newblock {Self-Refine: Iterative Refinement with Self-Feedback}.
\newblock In {\em NeurIPS}, 2023.

\bibitem[\protect\citeauthoryear{Mandi \bgroup \em et al.\egroup }{2024}]{zhao2024roco}
Zhao Mandi, Shreeya Jain, et~al.
\newblock {RoCo: Dialectic Multi-Robot Collaboration with Large Language Models}.
\newblock In {\em IEEE ICRA}, 2024.

\bibitem[\protect\citeauthoryear{Mirzadeh \bgroup \em et al.\egroup }{2024}]{mirzadeh2024gsmsymbolicunderstandinglimitationsmathematical}
Iman Mirzadeh, Keivan Alizadeh, et~al.
\newblock {GSM-Symbolic: Understanding the Limitations of Mathematical Reasoning in Large Language Models}, 2024.

\bibitem[\protect\citeauthoryear{Niwa and Iso}{2024}]{niwa-iso-2024-ambignlg}
Ayana Niwa and Hayate Iso.
\newblock {A}mbig{NLG}: Addressing task ambiguity in instruction for {NLG}.
\newblock In Yaser Al-Onaizan, Mohit Bansal, and Yun-Nung Chen, editors, {\em EMNLP 2024}, Miami, Florida, USA, November 2024. ACL.

\bibitem[\protect\citeauthoryear{Oelerich \bgroup \em et al.\egroup }{2024}]{oelerich2024languageguided}
Thies Oelerich, Christian Hartl-Nesic, et~al.
\newblock {Language-guided Manipulator Motion Planning with Bounded Task Space}.
\newblock In {\em CoRL}, 2024.

\bibitem[\protect\citeauthoryear{Oswald \bgroup \em et al.\egroup }{2024}]{oswald2024large}
James Oswald, Kavitha Srinivas, et~al.
\newblock {Large Language Models as Planning Domain Generators}.
\newblock In {\em ICAPS}, 2024.

\bibitem[\protect\citeauthoryear{Pearl and Mackenzie}{2018}]{bookofwhy}
Judea Pearl and Dana Mackenzie.
\newblock {\em {The Book of Why: The New Science of Cause and Effect}}.
\newblock Basic Books, Inc., 2018.

\bibitem[\protect\citeauthoryear{Puig \bgroup \em et al.\egroup }{2018}]{Puig_2018_virtualhome}
Xavier Puig, Kevin Ra, et~al.
\newblock {VirtualHome: Simulating Household Activities via Programs}.
\newblock In {\em IEEE CVPR}, 2018.

\bibitem[\protect\citeauthoryear{Richens and Everitt}{2024}]{richens2024robust}
Jonathan Richens and Tom Everitt.
\newblock {Robust Agents Learn Causal World Models}.
\newblock In {\em ICLR}, 2024.

\bibitem[\protect\citeauthoryear{Rocamonde \bgroup \em et al.\egroup }{2024}]{rocamonde2024visionlanguage}
Juan Rocamonde, Victoriano Montesinos, et~al.
\newblock {Vision-Language Models are Zero-Shot Reward Models for Reinforcement Learning}.
\newblock In {\em ICLR}, 2024.

\bibitem[\protect\citeauthoryear{Ruis \bgroup \em et al.\egroup }{2020}]{ruis2020gscan}
Laura Ruis, Jacob Andreas, et~al.
\newblock {A Benchmark for Systematic Generalization in Grounded Language Understanding}.
\newblock In {\em NeurIPS}, volume~33, 2020.

\bibitem[\protect\citeauthoryear{Saha \bgroup \em et al.\egroup }{2025}]{saha2025system}
Swarnadeep Saha, Archiki Prasad, et~al.
\newblock System 1.x: Learning to balance fast and slow planning with language models.
\newblock In {\em ICLR}, 2025.

\bibitem[\protect\citeauthoryear{Shen \bgroup \em et al.\egroup }{2024}]{shen2024taskbenchbenchmarkinglargelanguage}
Yongliang Shen, Kaitao Song, et~al.
\newblock {TaskBench: Benchmarking Large Language Models for Task Automation}, 2024.

\bibitem[\protect\citeauthoryear{Shinn \bgroup \em et al.\egroup }{2023}]{shinn2023reflexion}
Noah Shinn, Federico Cassano, et~al.
\newblock {Reflexion: language agents with verbal reinforcement learning}.
\newblock In {\em NeurIPS}, 2023.

\bibitem[\protect\citeauthoryear{Shridhar \bgroup \em et al.\egroup }{2021}]{ALFWorld20}
Mohit Shridhar, Xingdi Yuan, Marc-Alexandre C\^ot\'e, et~al.
\newblock {ALFWorld: Aligning Text and Embodied Environments for Interactive Learning}.
\newblock In {\em ICLR}, 2021.

\bibitem[\protect\citeauthoryear{Singh \bgroup \em et al.\egroup }{2022}]{singh2022progprompt}
Ishika Singh, Valts Blukis, et~al.
\newblock {ProgPrompt: Generating Situated Robot Task Plans using Large Language Models}.
\newblock In {\em Workshop on Language and Robotics at CoRL 2022}, 2022.

\bibitem[\protect\citeauthoryear{Smirnov \bgroup \em et al.\egroup }{2024}]{smirnov2024generatingconsistentpddldomains}
Pavel Smirnov, Frank Joublin, et~al.
\newblock Generating consistent pddl domains with large language models, 2024.

\bibitem[\protect\citeauthoryear{Song \bgroup \em et al.\egroup }{2024}]{song2024vlmsocialnavsociallyawarerobot}
Daeun Song, Jing Liang, et~al.
\newblock {VLM-Social-Nav: Socially Aware Robot Navigation through Scoring using Vision-Language Models}.
\newblock 2024.

\bibitem[\protect\citeauthoryear{Stechly \bgroup \em et al.\egroup }{2024a}]{stechly2024chainthoughtlessnessanalysiscot}
Kaya Stechly, Karthik Valmeekam, et~al.
\newblock {Chain of Thoughtlessness? An Analysis of CoT in Planning}.
\newblock {\em arXiv:2405.04776}, 2024.

\bibitem[\protect\citeauthoryear{Stechly \bgroup \em et al.\egroup }{2024b}]{stechly2024selfverificationlimitationslargelanguage}
Kaya Stechly, Karthik Valmeekam, et~al.
\newblock On the self-verification limitations of large language models on reasoning and planning tasks, 2024.

\bibitem[\protect\citeauthoryear{Sun \bgroup \em et al.\egroup }{2023}]{sun2023adaplanner}
Haotian Sun, Yuchen Zhuang, et~al.
\newblock {AdaPlanner: Adaptive Planning from Feedback with Language Models}.
\newblock In {\em NeurIPS}, 2023.

\bibitem[\protect\citeauthoryear{Tonmoy \bgroup \em et al.\egroup }{2024}]{tonmoy2024comprehensivesurveyhallucinationmitigation}
S.~M Towhidul~Islam Tonmoy, S~M~Mehedi Zaman, et~al.
\newblock A comprehensive survey of hallucination mitigation techniques in large language models, 2024.

\bibitem[\protect\citeauthoryear{Turpin \bgroup \em et al.\egroup }{2024}]{faithfulreasoning}
Miles Turpin, Julian Michael, et~al.
\newblock {Language models don't always say what they think: unfaithful explanations in chain-of-thought prompting}.
\newblock In {\em NeurIPS}, 2024.

\bibitem[\protect\citeauthoryear{Vafa \bgroup \em et al.\egroup }{2024}]{vafa2024implicitworldmodels}
Keyon Vafa, Justin~Y Chen, et~al.
\newblock {Evaluating the World Model Implicit in a Generative Model}.
\newblock {\em arXiv preprint arXiv:2406.03689}, 2024.

\bibitem[\protect\citeauthoryear{Valmeekam \bgroup \em et al.\egroup }{2023}]{valmeekam2023planbench}
Karthik Valmeekam, Matthew Marquez, et~al.
\newblock {PlanBench: An Extensible Benchmark for Evaluating Large Language Models on Planning and Reasoning about Change}.
\newblock In {\em NeurIPS Datasets and Benchmarks Track}, 2023.

\bibitem[\protect\citeauthoryear{Valmeekam \bgroup \em et al.\egroup }{2024}]{valmeekam2024lrms}
Karthik Valmeekam, Kaya Stechly, et~al.
\newblock {LLMs Still Can't Plan; Can LRMs? A Preliminary Evaluation of OpenAI's o1 on PlanBench}.
\newblock {\em arXiv}, 2024.

\bibitem[\protect\citeauthoryear{Verma \bgroup \em et al.\egroup }{2024}]{verma2024brittlefoundationsreactprompting}
Mudit Verma, Siddhant Bhambri, et~al.
\newblock {On the Brittle Foundations of ReAct Prompting for Agentic Large Language Models}, 2024.

\bibitem[\protect\citeauthoryear{Wang \bgroup \em et al.\egroup }{2023a}]{wang-etal-2023-plan}
Lei Wang, Wanyu Xu, et~al.
\newblock {Plan-and-Solve Prompting: Improving Zero-Shot Chain-of-Thought Reasoning by Large Language Models}.
\newblock In {\em Proceedings of ACL (Volume 1: Long Papers)}, 2023.

\bibitem[\protect\citeauthoryear{Wang \bgroup \em et al.\egroup }{2023b}]{wang2023selfconsistency}
Xuezhi Wang, Jason Wei, et~al.
\newblock {Self-Consistency Improves Chain of Thought Reasoning in Language Models}.
\newblock In {\em ICLR}, 2023.

\bibitem[\protect\citeauthoryear{Wang \bgroup \em et al.\egroup }{2024a}]{Wang_2024}
Lei Wang, Chen Ma, et~al.
\newblock {A survey on large language model based autonomous agents}.
\newblock {\em Frontiers of Computer Science}, 18(6), March 2024.

\bibitem[\protect\citeauthoryear{Wang \bgroup \em et al.\egroup }{2024b}]{wang2024llm}
Shu Wang, Muzhi Han, et~al.
\newblock {LLM}3: Large language model-based task and motion planning with motion failure reasoning.
\newblock In {\em Multi-modal Foundation Model meets Embodied AI Workshop @ ICML2024}, 2024.

\bibitem[\protect\citeauthoryear{Wang \bgroup \em et al.\egroup }{2024c}]{wang2024mint}
Xingyao Wang, Zihan Wang, et~al.
\newblock {MINT}: Evaluating {LLM}s in multi-turn interaction with tools and language feedback.
\newblock In {\em ICLR}, 2024.

\bibitem[\protect\citeauthoryear{Wei \bgroup \em et al.\egroup }{2022a}]{wei2022emergentabilitieslargelanguage}
Jason Wei, Yi~Tay, et~al.
\newblock {Emergent Abilities of Large Language Models}, 2022.

\bibitem[\protect\citeauthoryear{Wei \bgroup \em et al.\egroup }{2022b}]{wei2022chain}
Jason Wei, Xuezhi Wang, et~al.
\newblock {Chain of Thought Prompting Elicits Reasoning in Large Language Models}.
\newblock In {\em NeurIPS}, 2022.

\bibitem[\protect\citeauthoryear{Xie \bgroup \em et al.\egroup }{2023}]{xie2023translatingnaturallanguageplanning}
Yaqi Xie, Chen Yu, et~al.
\newblock {Translating Natural Language to Planning Goals with Large-Language Models}, 2023.

\bibitem[\protect\citeauthoryear{Xie \bgroup \em et al.\egroup }{2024a}]{xie2024travelplanner}
Jian Xie, Kai Zhang, et~al.
\newblock {TravelPlanner: A Benchmark for Real-World Planning with Language Agents}.
\newblock In {\em ICML}, 2024.

\bibitem[\protect\citeauthoryear{Xie \bgroup \em et al.\egroup }{2024b}]{xie2024osworld}
Tianbao Xie, Danyang Zhang, et~al.
\newblock {OSW}orld: Benchmarking multimodal agents for open-ended tasks in real computer environments.
\newblock In {\em NeurIPS Datasets and Benchmarks Track}, 2024.

\bibitem[\protect\citeauthoryear{Xie \bgroup \em et al.\egroup }{2024c}]{xie2024textreward}
Tianbao Xie, Siheng Zhao, et~al.
\newblock {Text2Reward: Reward Shaping with Language Models for Reinforcement Learning}.
\newblock In {\em ICLR}, 2024.

\bibitem[\protect\citeauthoryear{Yadav \bgroup \em et al.\egroup }{2021}]{review2021ambiguities}
Apurwa Yadav, Aarshil Patel, et~al.
\newblock A comprehensive review on resolving ambiguities in natural language processing.
\newblock {\em AI Open}, 2:85--92, 2021.

\bibitem[\protect\citeauthoryear{Yang \bgroup \em et al.\egroup }{2024}]{yang2024guidinglonghorizontaskmotion}
Zhutian Yang, Caelan Garrett, et~al.
\newblock {Guiding Long-Horizon Task and Motion Planning with Vision Language Models}.
\newblock In {\em CoRL 2024 Workshop on Language and Robot Learning Language as an Interface}, 2024.

\bibitem[\protect\citeauthoryear{Yao \bgroup \em et al.\egroup }{2022}]{yao22webshop}
Shunyu Yao, Howard Chen, et~al.
\newblock {WebShop: Towards Scalable Real-World Web Interaction with Grounded Language Agents}.
\newblock In {\em NeurIPS}, 2022.

\bibitem[\protect\citeauthoryear{Yao \bgroup \em et al.\egroup }{2023a}]{yao2023treethoughtsdeliberateproblem}
Shunyu Yao, Dian Yu, et~al.
\newblock {Tree of Thoughts: Deliberate Problem Solving with Large Language Models}.
\newblock In {\em NeurIPS}, 2023.

\bibitem[\protect\citeauthoryear{Yao \bgroup \em et al.\egroup }{2023b}]{yao2023tree}
Shunyu Yao, Dian Yu, et~al.
\newblock {Tree of Thoughts: Deliberate Problem Solving with Large Language Models}.
\newblock In {\em NeurIPS}, 2023.

\bibitem[\protect\citeauthoryear{Yao \bgroup \em et al.\egroup }{2023c}]{yao2023react}
Shunyu Yao, Jeffrey Zhao, et~al.
\newblock {{ReAct}: Synergizing Reasoning and Acting in Language Models}.
\newblock In {\em ICLR}, 2023.

\bibitem[\protect\citeauthoryear{Yildirim and Paul}{2024}]{whatdollmsknow}
Ilker Yildirim and L.A. Paul.
\newblock {From task structures to world models: what do LLMs know?}
\newblock {\em Trends in Cognitive Sciences}, 2024.

\bibitem[\protect\citeauthoryear{Yu \bgroup \em et al.\egroup }{2023}]{corlRewardLLM}
Wenhao Yu, Nimrod Gileadi, et~al.
\newblock {Language to Rewards for Robotic Skill Synthesis.}
\newblock In Jie Tan, Marc Toussaint, and Kourosh Darvish, editors, {\em CoRL}, Proceedings of Machine Learning Research. PMLR, 2023.

\bibitem[\protect\citeauthoryear{Yue \bgroup \em et al.\egroup }{2025}]{yue2025dots}
Murong Yue, Wenlin Yao, Haitao Mi, Dian Yu, Ziyu Yao, and Dong Yu.
\newblock {DOTS}: Learning to reason dynamically in {LLM}s via optimal reasoning trajectories search.
\newblock In {\em ICLR}, 2025.

\bibitem[\protect\citeauthoryear{Zhang \bgroup \em et al.\egroup }{2023}]{zhang2023groundingclassicaltaskplanners}
Xiaohan Zhang, Yan Ding, et~al.
\newblock {Grounding Classical Task Planners via Vision-Language Models}, 2023.

\bibitem[\protect\citeauthoryear{Zhang \bgroup \em et al.\egroup }{2024}]{zhang-etal-2024-pddlego}
Li~Zhang, Peter Jansen, Tianyi Zhang, Peter Clark, Chris Callison-Burch, and Niket Tandon.
\newblock {PDDLEGO}: Iterative planning in textual environments.
\newblock In {\em *SEM}, 2024.

\bibitem[\protect\citeauthoryear{Zhao \bgroup \em et al.\egroup }{2023}]{zhao2023llmmcts}
Zirui Zhao, Wee~Sun Lee, et~al.
\newblock {Large Language Models as Commonsense Knowledge for Large-Scale Task Planning}.
\newblock In {\em RSS 2023 Workshop on Learning for Task and Motion Planning}, 2023.

\bibitem[\protect\citeauthoryear{Zhong \bgroup \em et al.\egroup }{2024}]{zhong2024policy}
Victor Zhong, Dipendra Misra, et~al.
\newblock {Policy Improvement using Language Feedback Models}.
\newblock In {\em NeurIPS}, 2024.

\bibitem[\protect\citeauthoryear{Zhou \bgroup \em et al.\egroup }{2023}]{zhou2023leasttomost}
Denny Zhou, Nathanael Sch{\"a}rli, et~al.
\newblock {Least-to-Most Prompting Enables Complex Reasoning in Large Language Models}.
\newblock In {\em ICLR}, 2023.

\bibitem[\protect\citeauthoryear{Zhou \bgroup \em et al.\egroup }{2024}]{zhou2024webarena}
Shuyan Zhou, Frank~F. Xu, et~al.
\newblock {WebArena: A Realistic Web Environment for Building Autonomous Agents}.
\newblock In {\em ICLR}, 2024.

\end{thebibliography}


\begin{thebibliography}{}

\bibitem[\protect\citeauthoryear{Abelson \bgroup \em et al.\egroup
  }{1985}]{abelson-et-al:scheme}
Harold Abelson, Gerald~Jay Sussman, and Julie Sussman.
\newblock {\em Structure and Interpretation of Computer Programs}.
\newblock MIT Press, Cambridge, Massachusetts, 1985.

\bibitem[\protect\citeauthoryear{Baumgartner \bgroup \em et al.\egroup
  }{2001}]{bgf:Lixto}
Robert Baumgartner, Georg Gottlob, and Sergio Flesca.
\newblock Visual information extraction with {Lixto}.
\newblock In {\em Proceedings of the 27th International Conference on Very
  Large Databases}, pages 119--128, Rome, Italy, September 2001. Morgan
  Kaufmann.

\bibitem[\protect\citeauthoryear{Brachman and
  Schmolze}{1985}]{brachman-schmolze:kl-one}
Ronald~J. Brachman and James~G. Schmolze.
\newblock An overview of the {KL-ONE} knowledge representation system.
\newblock {\em Cognitive Science}, 9(2):171--216, April--June 1985.

\bibitem[\protect\citeauthoryear{Gottlob \bgroup \em et al.\egroup
  }{2002}]{gls:hypertrees}
Georg Gottlob, Nicola Leone, and Francesco Scarcello.
\newblock Hypertree decompositions and tractable queries.
\newblock {\em Journal of Computer and System Sciences}, 64(3):579--627, May
  2002.

\bibitem[\protect\citeauthoryear{Gottlob}{1992}]{gottlob:nonmon}
Georg Gottlob.
\newblock Complexity results for nonmonotonic logics.
\newblock {\em Journal of Logic and Computation}, 2(3):397--425, June 1992.

\bibitem[\protect\citeauthoryear{Levesque}{1984a}]{levesque:functional-foundations}
Hector~J. Levesque.
\newblock Foundations of a functional approach to knowledge representation.
\newblock {\em Artificial Intelligence}, 23(2):155--212, July 1984.

\bibitem[\protect\citeauthoryear{Levesque}{1984b}]{levesque:belief}
Hector~J. Levesque.
\newblock A logic of implicit and explicit belief.
\newblock In {\em Proceedings of the Fourth National Conference on Artificial
  Intelligence}, pages 198--202, Austin, Texas, August 1984. American
  Association for Artificial Intelligence.

\bibitem[\protect\citeauthoryear{Nebel}{2000}]{nebel:jair-2000}
Bernhard Nebel.
\newblock On the compilability and expressive power of propositional planning
  formalisms.
\newblock {\em Journal of Artificial Intelligence Research}, 12:271--315, 2000.

\end{thebibliography}

\end{document}